\documentclass[sigconf]{acmart}
\usepackage{graphicx}
\usepackage{xcolor}
\def\BibTeX{{\rm B\kern-.05em{\sc i\kern-.025em b}\kern-.08emT\kern-.1667em\lower.7ex\hbox{E}\kern-.125emX}}
    
\copyrightyear{2019}
\acmYear{2019}
\setcopyright{rightsretained}
\acmConference[KDD-DSHealth2019]{KDD ’19: The 25th ACM SIGKDD International Conference on Knowledge Discovery \& Data Mining}{Aug 5, 2019}{Anchorage, Alaska USA}
\acmBooktitle{2019 KDD Workshop on Applied Data Science for Healthcare (KDD-DSHealth2019), Aug 5, 2019, Anchorage, Alaska USA}
\acmPrice{}
\acmDOI{}
\acmISBN{}

\begin{document}

%
\title[Risk Adjustment in Claims using Medical Concept Representation Learning]{Medical Concept Representation Learning from Claims Data and Application to Health Plan Payment Risk Adjustment}
%
\author{Qiu-Yue Zhong}
\orcid{0000-0003-1666-8913}
\affiliation{%
	\institution{Geneia LLC}
	\streetaddress{50 Commercial St}
	\city{Manchester}
	\state{New Hampshire, USA}
	\postcode{03101}
}
\email{Qiuyue.Zhong@Geneia.com}

\author{Andrew H. Fairless}
\affiliation{%
	\institution{Geneia LLC}
	\streetaddress{50 Commercial St}
	\city{Manchester}
	\state{New Hampshire, USA}
	\postcode{03101}
}
\email{Andrew.Fairless@Geneia.com}

\author{Jasmine M. McCammon}
\affiliation{%
	\institution{Geneia LLC}
	\streetaddress{50 Commercial St}
	\city{Manchester}
	\state{New Hampshire, USA}
	\postcode{03101}
}
\email{Jasmine.McCammon@Geneia.com}

\author{Farbod Rahmanian}
\affiliation{%
	\institution{Geneia LLC}
	\streetaddress{1000 N Cameron St}
	\city{Harrisburg}
	\state{Pennsylvania, USA}
	\postcode{17103}
}
\email{Fred.Rahmanian@Geneia.com}

%
\renewcommand{\shortauthors}{Zhong, Fairless, McCammon and Rahmanian}

%
\begin{abstract}
Risk adjustment has become an increasingly important tool in healthcare. It has been extensively applied to payment adjustment for health plans to reflect the expected cost of providing coverage for members. Risk adjustment models are typically estimated using linear regression, which does not fully exploit the information in claims data. Moreover, the development of such linear regression models requires substantial domain expert knowledge and computational effort for data preprocessing. In this paper, we propose a novel approach for risk adjustment that uses semantic embeddings to represent patient medical histories. Embeddings efficiently represent medical concepts learned from diagnostic, procedure, and prescription codes in patients' medical histories. This approach substantially reduces the need for feature engineering. Our results show that models using embeddings had better performance than a commercial risk adjustment model on the task of prospective risk score prediction.  
\end{abstract}

%
%


\maketitle

\section{Introduction}
Risk adjustment is an important stabilization program in the health insurance market that aims to reduce incentives for avoiding costly patients. Without risk adjustment, plans may have incentives to enroll healthier patients and to avoid sick patients \cite{Schone2013-wx, Ellis2018-td, Hileman2016-dm}. Under risk adjustment, plans would receive higher payments for patients predicted to cost more (e.g., patients with multiple chronic conditions) and less for patients predicted to cost less (e.g., healthy patients). Hence, accurate risk adjustment models not only counteract avoidance of patients with greater health care needs but also provide an opportunity to provide care efficiently and generate profits by attracting the less healthy patients \cite{Schone2013-wx}.

Risk adjustment often begins with risk assessment, which assigns each patient a risk score that measures how costly that patient is expected to be \cite{Hileman2016-dm}. To calculate the risk scores, a formula is developed that predicts the cost as a function of patients' attributes such as demographic characteristics, prior expenditure, medical conditions extracted from diagnostic codes or medications, or self-reported health status \cite{Chang2010-yp, Schone2013-wx}. Typically, risk adjustment models are estimated using classical linear regression \cite{Rose2016-zz, Chang2010-yp}, which does not fully exploit the information in the data such as interactions and non-linear relationships between variables  \cite{Rose2016-zz}. In addition, although an increasing number of variables have been included in the models, they are unlikely to capture all of the factors that would affect expected costs \cite{Schone2013-wx}. Considering the current performance of popular risk adjustment models ($R^2$ ranged from 0.15 to 0.17) \cite{Schone2013-wx}, there is a substantial potential for performance improvement in risk adjustment models. Besides, developing risk adjustment models is a complicated process \cite{Ellis2018-td} which requires substantial feature engineering such as domain expert knowledge and data preprocessing.

In this paper, we propose to use semantic embeddings \cite{Le2014-ug,Mikolov2013-aw,Mikolov2013-jy} to extract patient representations for risk adjustment. Our work makes the following contributions:
\begin{itemize}
	\item We propose to use an established, easy-to-implement embedding algorithm to learn generic patient-level representations in claims data without depending on expert medical knowledge and heavy data preprocessing. 
	\item We apply the learned representations to predicting prospective risk scores and demonstrate the superior performance of the proposed semantic embeddings compared to a commercial risk adjustment model.
	\item We use both linear and non-linear machine learning algorithms for predicting prospective risk scores and show the performance improvement of non-linear algorithms.
\end{itemize}

\section{Related Work}
Early statistical language models typically used schemes, such as one-hot encoding, that represented words as discrete, independent units. This sparse representation was inefficient to store and process. It also hampered the ability of these models to generalize patterns learned from one set of words to other combinations of words \cite{Beam2018-ap, Bengio2003-re}. 

Distributed representations of words solve both of these problems. First, they more efficiently represent words as dense vectors. Second, they can encode semantic and syntactic similarities and differences among words by arranging similar words near each other in the embedded vector space (i.e., semantic embeddings) and different words far apart. For example, \textit{cat} and \textit{dog} might be placed near each other, whereas both might be far from \textit{blue} or \textit{write}. Based on the similarity between \textit{cat} and \textit{dog}, a language model that is trained on word vectors and has encountered the sentence \textit{"The cat runs."} might more readily generate the grammatically and semantically sensible sentence \textit{"The dog runs."} \cite{Bengio2003-re, Mikolov2013-jy}

Historically, the major obstacle to using high-quality word vectors was creating them within a feasible computational time. Mikolov et al. \cite{Mikolov2013-jy, Mikolov2013-aw} largely overcame this obstacle by proposing the \textit{word2vec} models, the continuous bag-of-words (CBOW) model and the continuous skip-gram model. Both models work by repeatedly selecting a target word and a specified number of surrounding context words. The CBOW model trains to predict the target word based on the context words' vectors, while the skip-gram model predicts the context words from the target word's vector. As successive sets of target and context words are used for training, the embedded word vectors are adjusted to minimize the prediction error so that high-quality embeddings are learned.  

Le and Mikolov \cite{Le2014-ug} extended the \textit{word2vec} models to groups of words, including sentences, paragraphs, and entire documents. In their Distributed Memory Model of Paragraph Vectors (PV-DM), which is analogous to the CBOW model, a paragraph (or other chosen word group) vector is added as a predictor to the context words' vectors. The paragraph vector "remembers" information about the paragraph beyond the selected context words and thus helps to predict the target word. The Distributed Bag of Words version of Paragraph Vectors (PV-DBOW) is analogous to the skip-gram model and uses only the paragraph vector to predict context words from the same paragraph.  PV-DM provides the advantage of accounting for the sequence of the words in the paragraph, while PV-DBOW is less computationally intensive. Both, which are collectively known as \textit{doc2vec}, allow efficient learning of paragraph vectors, even though different paragraphs may vary in length.

Given the success in modeling language, embeddings have been adapted to other domains, including healthcare. Researchers have used embeddings created from electronic health records (EHRs), claims data, and published biomedical literature to represent biomedical concepts, billing codes, patient visits to providers, and patients' medical histories \cite{Beam2018-ap, Rajkomar2018-zh, Miotto2016-lw, Choi2016-ll, Bajor2018-wj, Pham2017-ie, Zhang2018-ph, Vine2014-qj, Choi2016-dv}. The last one, medical histories, are especially complex to model without embeddings because they subsume the other three; a complete and accurate representation would require modeling the similarities and differences among biomedical concepts, how they map to medical codes, and the irregular time intervals between clinical visits. Traditionally, such modeling would require slow, expensive manual crafting by biomedical and statistical experts. Miotto et al. \cite{Miotto2016-lw} and Bajor et al. \cite{Bajor2018-wj} circumvented this slow process by using autoencoders and the \textit{doc2vec} models, respectively, to encode medical histories from EHRs. Our approach is most similar to that of Bajor et al. \cite{Bajor2018-wj}, except that we create our embeddings from claims data, instead of from EHRs.

\section{Patient-level Embeddings}

\subsection{Source of Data}
We used de-identified claims data provided by a medium-sized, regional U.S. health insurance plan over the period of 2015-2016. The data contains demographic and enrollment information, cost, diagnostic and procedure codes, and medications. We included patients who had (1) at least one month enrollment in both years; (2) at least one medical or pharmacy claims in both years; and (3) risk scores in 2016 from a commercial risk adjustment model. A total of 441,271 patients were included in this study. 

To train embeddings, we extracted all diagnoses (International Classification of Diseases [ICD]-9 or ICD-10), procedure (Current Procedural Terminology [CPT]) codes, and medications (National Drug Code [NDC]), along with their associated date stamps from each patient's record. No data preprocessing was involved: these codes were used as-is. For each patient, we ordered the codes chronologically. 

\subsection{Embedding Training}
We treated a patient's entire record as a "document," and codes in the patient's record as "words" in the document. For each patient's record, we computed an embedded representation using both \textit{doc2vec} models (PV-DBOW and PV-DM). For each model, we trained combinations of embedding dimensions (100, 200, 300), sliding window sizes (10, 15, 20) with negative sampling. We used 159,457,590 codes from 441,271 patients for embedding training. All models were generated using Gensim in Python.

\section{Experiments}
We conducted all our experiments using the same data source as the one used for embedding training.

\subsection{Prediction Task}
The prediction task was to predict prospective risk scores in 2016 using information from 2015. To calculate the risk scores in 2016, we first summed the total allowed costs in 2016. We then weighted each patient's total cost by the total length of enrollment in 2016. Lastly, we rescaled the weighted costs to have a mean of 1.0 over the population \cite{Hileman2007-dk}.

\subsection{Baseline Models}
\subsubsection{Baseline Models 1} We developed our own risk adjustment model. We designed 21 features (Table 1) from 2015 data. These included age, sex, clinical characteristics measured using diagnostic and procedure codes, medications, healthcare utilization, and total cost. We also included a community-level description of race based on each patient' residential ZIP code.

\subsubsection{Baseline Models 2} 
We used a commercial risk adjustment model as another baseline model. This simple linear-additive model includes age, sex, and $\sim$150 medical markers and pharmacy markers from 2015 as features to calculate risk scores for 2016 \cite{Hileman2007-dk}.

\subsection{Embeddings-based Model} 
We used the trained embedded representations as input features for the prediction task.

\subsection{Experimental Design}
We evaluated the embedded representations by assessing how well they perform as input features for predicting risk scores in 2016, comparing them to the performance of the baseline models.

For the embeddings-based models, we performed a grid search over parameters (model, embedding dimension, and window size) using RIDGE regression optimized by cross-validation. We chose the best embedding parameter setting (PV-DBOW with an embedding dimension of 100 and a window size of 15) and performed the subsequent experiments using this best parameter setting.

We implemented our experiments for Baseline Model 1 and the embeddings-based model using different algorithms including RIDGE regression and extreme gradient boosting (XGBoost) \cite{Chen2016-rw}. 

To be consistent with previous studies in risk adjustment \cite{Hileman2007-dk}, we computed three measures to evaluate the predictive accuracy: (1) $R^2$; (2) mean absolute error (MAE); and (3) predictive ratios (PRs). A predictive ratio evaluates the predictive fit at a group level and is defined as the mean predicted risk score divided by the mean actual cost for a subgroup of individuals from the sample population, with both values scaled to 1.0 over the entire population. A PR closest to 1.0 indicate a very good fit for a particular group. A predictive ratio$>1$ indicates that a model overestimates the risk level for that group, whereas a PR $<1$ indicates that the model underestimates the risk level. We calculated predictive ratios by age and sex groups. 

We implemented all the experiments with Python. We split the dataset into training $(70 \%)$ and test $(30 \%)$ sets.  

\subsection{Prediction Results}
Table 1 summarizes the patients' characteristics in the training and test sets.

\begin{table}
	\caption{Characteristics of Patients. Cells contain Median [P25, P75] if not otherwise stated.}
	\scalebox{0.73}{
	\begin{tabular}{lcc}
		\toprule
		\textbf{Characteristics} &\textbf{Training Set}&\textbf{Test Set}\\
		&$N= 308,889$& $N = 132,382$\\
		\midrule
		Age in 2015, year&41.0 [20.0, 56.0]&41.0 [20.0, 56.0]\\
		Female, n (\%)&167,271 (54.2)&71,886 (54.3)\\
		\% of population that is African American&3.4 [2.7, 7.5]&3.4 [2.7, 7.5]\\ 
		(based on 3-digit ZIP codes)&&\\
		Charlson Comorbidity Index&0 [0, 1]&0 [0, 1]\\
		\# of claims for inpatient visits in 2015&0 [0, 0]&0 [0, 0]\\
		\# of claims for outpatient visits in 2015&6 [3, 13]&6 [3, 13]\\	
		\# of claims for emergency department &0 [0, 0]&0 [0, 0]\\
		visits in 2015&&\\
		\# of pharmacy claims in 2015&3 [0, 19]& 3 [0, 20]\\
		\# of claims for specialty drugs in 2015&0 [0, 0]&0 [0, 0]\\
		\# of distinct drug classes in 2015&1 [0, 4]&1 [0, 4]\\
		Chemotherapy in 2015, n (\%)&40,936 (13.3)&17,621 (13.3)\\
		Psychotherapy in 2015, n (\%)&24,323	(7.9)&10,396 (7.9)\\
		Obesity in 2015, n (\%)&59,010 (19.1)&25,395	(19.2)\\	
		Cardiovascular disease in 2015, n (\%)&29,002	(9.5)&12,531 (9.5)\\
		Hypertension in 2015, n (\%)&83,836 	(27.1)&36,197 (27.3)\\
		Type II diabetes in 2015, n (\%)&28,424	(9.2)&12,334 (9.3)\\
		Mental disorders in 2015, n (\%)&106,797	(34.6)&46,029 (34.8)\\	
		Drug/alcohol abuse in 2015, n (\%)&7,916	(2.6)&3,385 (2.6)\\
		Low back pain in 2015, n (\%)&47,587	(15.4)&20,467	(15.5)\\
		Asthma in 2015, n (\%)&32,414 (10.5)&13,826	(10.4)\\
		Cost in 2015&1,591.1 [547.9, 4,909.0]&1,601.5	[555.4, 4,961.1]\\
		Cost in 2016&1,634.6 [556.2, 5,133.3]&1,636.8	[556.3, 5,087.3]\\			
		\bottomrule
	\end{tabular}}
\end{table}

Table 2 shows the experimental results for individual-level predictive fit of three different approaches. The embeddings-based model greatly outperformed Baseline Model 2 and had a comparable performance to Baseline Model 1. For both Baseline Model 1 and the embeddings-based model, non-linear algorithms (XGBoost) had better performance than simple, linear algorithms (RIDGE). 

\begin{table}
	\caption{$R^2$ and Mean Absolute Error (MAE) of Baseline Models and Embeddings-based Model in the Test Set ($N = 132,382$)}
	\begin{tabular}{lrr}
		\toprule
		\textbf{Models} &$\mathbf{R^2}$& \textbf{MAE}\\
		\midrule
		\textbf{Baseline Model 1} & & \\
		\hspace{3mm} RIDGE & 0.41& 0.88\\
		\hspace{3mm} XGBoost & 0.52& 0.72\\
		\textbf{Baseline Model 2} & 0.04 & 1.07\\
		\textbf{Embeddings-based Model}&&\\
		\hspace{3mm} RIDGE & 0.40& 0.84\\
		\hspace{3mm} XGBoost & 0.54& 0.73\\	
		\bottomrule
	\end{tabular}
\end{table} 

Table 3 demonstrates the predictive ratios for group-level fit. Due to space limitation, we only present results from XGBoost. The embeddings-based model overestimated the risk for the elderly population with more evident bias among female and underestimated the risk for children aged 0-2. Overall, the embeddings-based model showed less evident bias as compared to Baseline Model 2.

\begin{table}
	\caption{Predictive Ratios (PRs) by Age and Sex ($N = 441,271$)}
	\scalebox{0.665}{
	\begin{tabular}{lrrccc}
		\toprule
		\textbf{Sex} & \textbf{Age} & $\mathbf{N}$ &&\textbf{PRs}&\\
		&&&	\textbf{Baseline Model 1}&\textbf{Baseline Model 2}&\textbf{Embeddings-based Model}\\
		&&&\textbf{XGBoost}&&\textbf{XGBoost}\\
		\midrule
	    Male& (0, 1]&2,394&1.080&0.945&0.840\\
	    Male& (1, 2]&2,528&1.159&0.824&0.975\\
	    Male& (2, 4]&5,234&1.182&0.812&1.054\\
	    Male& (4, 9]&13,892&1.394&0.707&1.126\\
	    Male& (9, 14]&15,176&1.235&0.765&0.978\\
	    Male& (14, 18]&12,771&1.035&0.634&0.917\\
	    Male& (18, 20]&5,084&1.126&0.658&1.018\\
	    Male& (20, 24]&8,751&1.077&0.646&1.105\\
	   	Male& (24, 29]&8,821&1.018&0.636&1.027\\
	   	Male& (29, 34]&9,848&1.091&0.691&1.116\\
	   	Male& (34, 39]&11,217&1.096&0.764&1.130\\
	   	Male& (39, 44]&12,529&1.053&0.783&1.066\\
	   	Male& (44, 49]&15,752&1.002&0.851&1.002\\
	   	Male& (49, 54]&18,199&0.964&0.824&0.950\\
	   	Male& (54, 59]&20,539&0.911&0.832&0.890\\
	    Male& (59, 64]&19,807&0.885&0.871&0.867\\
	   	Male& (64, 69]&8,995 &0.884&1.118&0.889\\
	    Male& (69, 74]&4,439&0.957&1.887&1.052\\
	    Male& (74, 79]&2,418&1.005&2.211&1.169\\
	    Male& (79, 84]&1,924&0.975&2.301&1.196\\
	    Male& 84+ &1,795&0.950&2.389&1.289\\
	    Female& (0, 1]&2,268&1.093&0.933&0.857\\
		Female& (1, 2]&2,489&1.110&0.727&0.963\\
		Female& (2, 4]&4,943&1.315&0.868&1.164\\
		Female& (4, 9]&12,850&1.406&0.666&1.150\\
		Female& (9, 14]&14,238&1.236&0.862&1.027\\
		Female& (14, 18]&13,259&1.127&0.728&1.017\\
		Female& (18, 20]&6,511&1.050&0.640&0.956\\
		Female& (20, 24]&13,089&1.053&0.724&1.071\\
		Female& (24, 29]&14,600&1.022&0.820&1.061\\
		Female& (29, 34]&14,804&0.993&0.761&1.054\\
		Female& (34, 39]&15,676&1.031&0.749&1.033\\
		Female& (39, 44]&16,183&1.019&0.740&1.041\\
		Female& (44, 49]&19,363&1.048&0.784&1.023\\
		Female& (49, 54]&21,803&0.956&0.737&0.930\\
		Female& (54, 59]&23,523&0.991&0.811&0.963\\
		Female& (59, 64]&21,299&0.943&0.857&0.916\\
		Female& (64, 69]&8,792&0.949&1.214&0.971\\
		Female& (69, 74]&4,786&1.020&2.379&1.212\\
		Female& (74, 79]&2,784&1.040&2.680&1.274\\
		Female& (79, 84]&2,625&1.064&2.877&1.299\\
		Female& 84+ &3,271&1.041&2.853&1.277\\
		\bottomrule
	\end{tabular}}
\end{table}

\section{Conclusions}
In this paper, we proposed to use patient-level semantic embeddings for plan payment risk adjustment. In the experiments, the embeddings-based model showed improved performance in both individual- and group-level predictive fit compared to a commercial risk adjustment tool. Our method provided a rapid and easy-to-implement approach for risk adjustment that did not heavily rely on domain expertise, nor did it require extensive data preprocessing. A key advantage of our method was that we did not need to specify which combinations of which potential features to include in the model; instead, our approach was able to learn representations of the key factors and interactions from the claims data itself. Additionally, our method may reduce opportunities for gaming the risk adjustment system \cite{Ellis2018-td, Rose2016-zz}, such as "upcoding" where extra diagnostic codes or codes representing the most costly condition that a member might have are recorded \cite{Schone2013-wx}. We also showed that the non-linear algorithm predicted risk scores more accurately than a linear approach. Considering the large number of individuals enrolled in insurance programs using risk adjustment, the cost-saving implications of improved risk adjustment are immense \cite{Rose2016-zz}. Although our approach to risk adjustment requires further empirical evaluation, our initial work with machine learning techniques is promising.  

Significant prior research has focused on developing patient representations from EHRs. However, very few studies \cite{Choi2016-dv} have primarily focused on claims data, which are widely available to both providers and payers. Unlike EHRs, claims data does not include clinical notes, laboratory test results, or vital signs. However, it does encompass every interaction that a patient has with any healthcare provider, as long as a billing claim is generated. We have demonstrated the effectiveness of semantic embeddings developed solely based on claims data in a common prediction task in the health insurance domain. Given the generic nature of the embedded representation, our approach can be easily applied to a wide variety of prediction problems. In addition, future work using claims data can consider using pre-trained distributed representations learned from multimodal medical data \cite{Beam2018-ap}, such as using claims, EHRs, and medical journals, to see if such representations will further improve the performance.


\bibliographystyle{ACM-Reference-Format}
\bibliography{ref_emb_20190405}
\end{document}